\documentclass{article}

\usepackage{latexml}

\usepackage[eandd, final]{neurips_2026}

\usepackage[utf8]{inputenc} 
\usepackage[T1]{fontenc}    
\usepackage{url}            
\usepackage{booktabs}       
\usepackage{colortbl}       
\usepackage{array}          
\usepackage{makecell}       
\usepackage{amsfonts}       
\usepackage{amsmath}        
\usepackage{nicefrac}       
\usepackage{textcomp}       
\usepackage{microtype}      
\usepackage{xcolor}         
\usepackage{graphicx}       
\usepackage{wrapfig}        

\usepackage{pifont}

\usepackage{amsmath,amsfonts,bm}

\def\eqref#1{equation~\ref{#1}}

\def\1{\bm{1}}

\def\vw{{\bm{w}}}
\def\vx{{\bm{x}}}
\def\vy{{\bm{y}}}
\def\vz{{\bm{z}}}

\def\mW{{\bm{W}}}
\def\mX{{\bm{X}}}
\def\mY{{\bm{Y}}}
\def\mZ{{\bm{Z}}}

\DeclareMathAlphabet{\mathsfit}{\encodingdefault}{\sfdefault}{m}{sl}
\SetMathAlphabet{\mathsfit}{bold}{\encodingdefault}{\sfdefault}{bx}{n}

\usepackage{enumitem}
\usepackage{tailwind_colors}
\usepackage{mdframed}
\arrayrulecolor{slate300}   
\usepackage{amssymb}
\iflatexml
  \newcommand{\papernum}[2]{#2}
\else
  \usepackage[group-separator={,},output-decimal-marker={.},mode=text]{siunitx}
  \newcommand{\papernum}[2]{\num{#1}}
\fi

\usepackage[colorlinks=true,linkcolor=blue600,citecolor=blue600,urlcolor=indigo600]{hyperref}

\usepackage{changepage}
\usepackage{xspace}
\usepackage{multirow}
\usepackage{listings}

\iflatexml
  \newcommand{\cmark}{Yes}
  \newcommand{\xmark}{No}
\else
  \usepackage{fontawesome}
  \newcommand{\cmark}{{\small\color{indigo500}\faCheckCircle}}
  \newcommand{\xmark}{{\small\color{slate300}\faTimesCircle}}
\fi

\iflatexml\else
\fi

\newcommand{\dataset}{\texttt{Cephalonauts One}\xspace}
\lstdefinestyle{pythoncode}{
  language=Python,
  basicstyle=\ttfamily\small,
  backgroundcolor=\color{slate50},
  commentstyle=\color{gray600},
  keywordstyle=\color{indigo600}\bfseries,
  rulecolor=\color{indigo300},
  stringstyle=\color{emerald600},
  breaklines=true,
  showstringspaces=false,
  columns=fullflexible,
  keepspaces=true,
  frame=leftline,
  framerule=1.2pt,
  xleftmargin=5pt,
  xrightmargin=0pt,
  framexleftmargin=1pt,
  aboveskip=0.75em,
  belowskip=0.5em
}

\iflatexml
\title{Cephalonauts One: A deep fMRI dataset for decoding naturalistic speech in the human brain}
\else
\title{
  \dataset: A deep fMRI dataset\\
  for decoding naturalistic speech in the human brain
}
\fi

\iflatexml
\author{%
  Antoine Collas\\ Karavela\\ \texttt{acollas@karavela.ai}
  \and Louis Jalouzot\\ Karavela; UNICOG, CNRS, INSERM, CEA, Paris-Saclay University; LSCP, EHESS, ENS, CNRS, PSL University
  \and Géraud Ilinca\\ Karavela
  \and Corentin Caris\\ Karavela
  \and Romain Valabrègue\\ Centre de NeuroImagerie de Recherche (CENIR), Sorbonne Université, ICM, Paris, France
  \and Ahmed Hassayoune\\ Karavela
  \and David Goncalves\\ Karavela
  \and Madeleine Hueber\\ Karavela
  \and Thaddée Delebarre\\ Karavela
  \and Julien Savatovsky\\ Department of Radiology, Hôpital Fondation Adolphe de Rothschild, Paris, France
  \and Clara Fonteneau\\ Karavela
  \and Charles Maussion\\ Karavela
  \and Bertrand Thirion\\ Karavela; Inria, CEA, Paris-Saclay University, Palaiseau, France\\ \texttt{bthirion@karavela.ai}
  \and Alexis Thual\\ Karavela\\ \texttt{athual@karavela.ai}
}
\else
\author{%
    \textbf{Antoine Collas$^{\dagger}$,
    Louis Jalouzot$^{\dagger,\circ,\diamond}$,
    Géraud Ilinca$^{\dagger}$,
    Corentin Caris$^{\dagger}$,
    Romain Valabrègue$^{\ddagger}$,} \\
    \textbf{Ahmed Hassayoune$^{\dagger}$,
    David Goncalves$^{\dagger}$,
    Madeleine Hueber$^{\dagger}$,
    Thaddée Delebarre$^{\dagger}$,} \\
    \textbf{Julien Savatovsky$^{\mathsection}$,
    Clara Fonteneau$^{\dagger}$,
    Charles Maussion$^{\dagger}$,
    Bertrand Thirion$^{\dagger,\P}$,
    Alexis Thual$^{\dagger}$}\\\\
    $^{\dagger}$Karavela \\
    $^{\circ}$UNICOG, CNRS, INSERM, CEA, Paris-Saclay University \\
    $^{\diamond}$LSCP, EHESS, ENS, CNRS, PSL University \\
    $^{\ddagger}$Centre de NeuroImagerie de Recherche (CENIR), Sorbonne Université, ICM, Paris, France \\
    $^{\mathsection}$Department of Radiology, Hôpital Fondation Adolphe de Rothschild, Paris, France \\
    $^{\P}$Inria, CEA, Paris-Saclay University, Palaiseau, France \\\\
    \texttt{\{acollas, athual, bthirion\}\ @\ karavela.ai}
}

\fi

\begin{document}
\maketitle

\begin{abstract}
    \dataset is a whole-brain 3 Tesla (3T) functional magnetic resonance imaging (fMRI) dataset recorded while subjects listened to audio podcasts.
    Three healthy subjects underwent multiple scanning sessions, each consisting of five 15-minute runs, while listening to podcasts in their native language.
    With 30 hours of fMRI data per subject, the current release is the deepest available fMRI dataset using naturalistic speech stimuli.
    The dataset pairs brain activity with the corresponding podcast audio, transcript annotations, and derived stimulus embeddings.
    Furthermore, we introduce a brain decoding benchmark formulated as audio segment retrieval: given fMRI activity from a held-out session,
    the decoder must identify the corresponding time-aligned podcast audio segment among candidate segments.
    We provide standardized splits, evaluation metrics, and baseline decoders for this task.
    Finally, a scaling analysis shows that decoding performance improves continuously with the amount of training data per subject.
    
    \noindent\begin{tabular}{@{}l@{\quad}l@{}}
    \textbf{Data \& code:} & {\color{indigo600}\texttt{\href{https://huggingface.co/datasets/karavela/cephalonauts_one}{huggingface.co/datasets/karavela/cephalonauts\_one}}} \\
    \textbf{Demo:} & {\color{indigo600}\texttt{\href{https://karavela.ai/en/research/cephalonauts-one}{karavela.ai/en/research/cephalonauts-one}}}
    \end{tabular}
\end{abstract}

\begin{figure}[h]
  \centering
  \iflatexml
    \includegraphics[width=800pt]{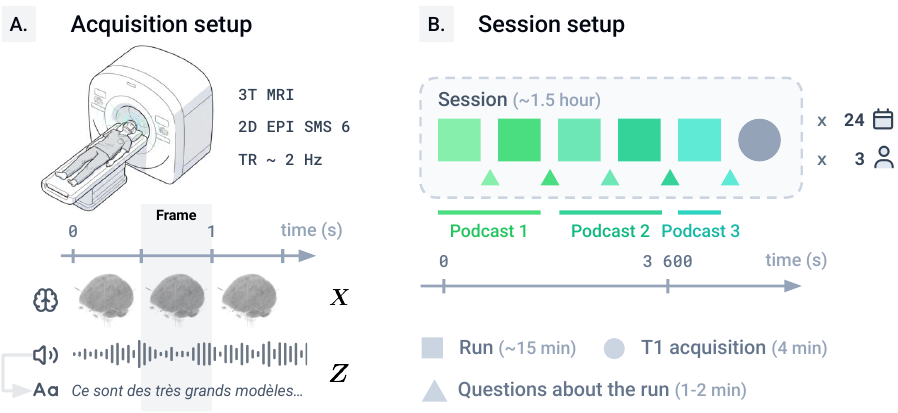}
  \else
    \makebox[\textwidth][c]{\includegraphics[width=0.9\textwidth]{figures/intro.pdf}}
  \fi
  \caption{
    \textbf{Overview of the \dataset dataset.}
    \textbf{A.} Three subjects listened to podcasts during fMRI acquisition.
    \textbf{B.} Each session included five 15-minute runs, three comprehension questions after each run, and a final anatomical scan. Runs 1--2, 3--4, and 5 used three distinct podcasts and podcasts were not repeated across sessions.
  }
  \label{fig:dataset_overview}
\end{figure}

\section{Introduction}
\textbf{Brain-behavior links}
The human brain can process highly complex perceptual inputs but also produce rich, intentional, and reflective thoughts.
Extensive research has been devoted to probing, measuring, and modelling this complex biological system, in the hope of gaining clinical insights into brain disorders, inspiring the design of more intelligent artificial systems, and better grasping the nature of human cognition~\citep{mesulam1998sensation}.
One way of gaining information about the brain is to link its properties -- e.g., anatomical, functional, or biological -- to properties of the external world -- e.g., a list of tasks an individual is able or unable to perform, or the category of stimuli they are exposed to.
Famous historical approaches demonstrated that lesions in specific frontal and temporal regions could impair speech and comprehension, yielding early insights into the functional organization of the language system~\citep{broca_perte_1865,wernicke_aphasische_1874}.

\textbf{Brain encoding and decoding}
More recently, functional magnetic resonance imaging (fMRI)~\citep{huettel2004functional} enabled non-invasive measurement of brain activity in healthy subjects.
Its higher spatial and temporal resolution compared with positron emission tomography (PET) made it possible to characterize brain responses in greater detail.
This led to early statistical analyses showing that some brain regions respond selectively to specific stimulus categories~\citep{kanwisher_fusiform_1997}.
To identify regions where these effects cannot necessarily be observed so clearly, a large body of work has predicted recorded brain activity from known stimulus features, a process known as \textit{encoding}.
Encoding models characterize how stimulus features are represented across the cortex, from retinotopic organization in early visual areas \citep{dumoulin2008population} to semantic maps estimated from fMRI recorded while participants listened to long natural narratives \citep{huth_natural_2016}.
In parallel, many studies have attempted to infer properties of the stimulus from recorded brain activity, a process known as \textit{decoding}.
In language neuroscience, recent decoding approaches have moved beyond coarse stimulus labels by mapping non-invasive brain recordings to rich speech representations.
\citet{jalouzot_optimizing_2025} showed that fMRI decoding of natural speech improves with additional subject-specific data, a finding consistent with related speech-decoding results in magneto- and electroencephalography (M/EEG)~\citep{defossez_decoding_2023}.

\textbf{Stimulus embeddings}
In fact, brain encoding and decoding can be viewed as two sides of the same coin, as both aim to link properties of brain activity to properties of the stimulus.
While early work focused on relatively low-dimensional variables such as stimulus category or the presence of cognitive impairments, the advent of deep learning has made it possible to access high-level high-dimensional representations of the stimulus---so-called \textit{latent representations} or \textit{embeddings}---that can be used in both encoding \citep{eickenberg_seeing_2017} and decoding analyses \citep{defossez_decoding_2023}.
In both settings, these representations support increasingly expressive models with greater predictive power \citep{dascoli_foundation_2026}, opening new perspectives for understanding brain function.

\textbf{Natural speech}
Language is a particularly compelling stimulus, spanning multiple levels from acoustics to semantics and narrative context.
Spoken narratives are especially useful for studying these representations in the brain: they engage distributed cortical systems~\citep{huth_natural_2016} over multiple timescales~\citep{lerner2011topographic}, while remaining easy to align with transcripts and computational representations from modern \textit{large language models} (LLMs)~\citep{zhang_qwen3_2025,khare2025universal}.
Prior work used brain responses to compare natural language processing embeddings, showing that they capture different aspects of language~\citep{toneva_interpreting_2019}.
Other studies have shown that modern models partly match human brain responses~\citep{caucheteux_brains_2022}, or used controlled models to separate the effects of meaning, syntax, and context in the cortex~\citep{pasquiou_information_2023}.
Together, these results show that fMRI during speech listening can help study both biological and computational language representations.

\textbf{Dataset depth}
However, fMRI with speech stimuli remains constrained by limited per-participant data.
This is limiting for subject-specific decoders, for which performance improves with additional within-subject data in both natural speech and visual fMRI decoding~\citep{tang_semantic_2023,scotti_reconstructing_2023}.
Indeed, large-scale analyses further indicate that decoding scales primarily with data per subject~\citep{banville_scaling_2025}.
Existing speech fMRI resources address different regimes: Narratives provides up to roughly 2 hours of unique stimuli per participant~\citep{nastase_narratives_2021}, whereas \citet{lebel_natural_2023} provide up to 16 hours for 3 participants.
Hence, deep per-subject language fMRI datasets for natural speech encoding and decoding remain scarce.

\section{Related work}

\begin{table}[t]
  \caption{
    \textbf{Brain datasets using speech stimuli.}
    \textit{Maximum hours per subject} reports the maximum subject-specific stimulus duration.
    For \citet{lebel_natural_2023} we report the 2 available data regimes.
    \textit{Benchmark-oriented} denotes releases with benchmark tasks, metrics, baselines, and code.
  }
  \label{table:related_datasets}
  \renewcommand{\arraystretch}{1.1}
  \begin{adjustwidth}{-1cm}{-1cm}
  \centering
  \setlength{\tabcolsep}{4pt}
  \begin{tabular}{lccccc}
    \hline
    \rowcolor{slate100}
    \textbf{Dataset} &
    \textbf{Modality} &
    \makecell{\textbf{Nb. of}\\\textbf{subjects}} &
    \makecell{\textbf{Max hours}\\\textbf{per subject}} &
    \makecell{\textbf{Benchmark-}\\\textbf{oriented}} &
    \textbf{References} \\
    \hline
    \dataset & fMRI & 3 & 30 & \cmark & This work \\
    LibriBrain & MEG & 1 & 52.3 & \cmark & \citealp{ozdogan2025libribrain} \\
    LeBel et al. & fMRI & 8 / 3 & 6 / 16 & \xmark & \citealp{lebel_natural_2023} \\
    Le Petit Prince & fMRI & 112 & 1.7 & \xmark & \citealp{li_petit_prince_2022} \\
    Armeni et al. & MEG & 3 & 10 & \xmark & \citealp{armeni_2022} \\
    Narratives & fMRI & 345 & 2 & \xmark & \citealp{nastase_narratives_2021} \\
    \hline
  \end{tabular}
  \end{adjustwidth}
\end{table}

\textbf{Early scaling of naturalistic fMRI datasets}
Public neural datasets have been central to progress in brain encoding and decoding.
Early scaling was driven by fMRI datasets based on visual stimuli.
Indeed, early decoding datasets contained roughly 1.5--3 stimulus hours per subject~\citep{haiguang_wen_data_2017,chang_bold5000_2019}.
Then, the Natural Scenes Dataset increased visual stimuli to 40 hours per subject~\citep{allen_massive_2022}.
More recently, the Courtois NeuroMod datasets pushed further with roughly 80 hours of movie watching per subject~\citep{boyle_cneuromod_2025}.
This scale has supported advances in visual encoding and decoding, including natural-image reconstruction from fMRI with deep learning models~\citep{chen_seeing_2022,ozcelik_natural_2023,scotti_reconstructing_2023}.

\textbf{Speech stimuli in fMRI datasets: breadth and depth}
However, speech fMRI datasets have often scaled across rather than within participants.
Narratives aggregates story-listening fMRI from 345 participants, with roughly 4.6 hours of unique audio stimuli across the full dataset~\citep{nastase_narratives_2021}, while Le Petit Prince provides multilingual audiobook fMRI from 112 subjects, with approximately 1.7 hours per subject~\citep{li_petit_prince_2022}.
Shared stimuli support group analyses and cross-subject comparisons, but limited per-subject depth constrains subject-specific decoders.
Recently, \citet{lebel_natural_2023} went deeper with natural story-listening fMRI from 8 participants, including an extended subset of up to 16 hours of podcast listening for 3 participants.
Using these data, \citet{jalouzot_optimizing_2025} show that natural speech decoding improves with more training data per participant.

\textbf{Benchmark-ready datasets}
In an effort to make brain encoding and decoding more reproducible and accessible, recent datasets provide standardized tasks, code, and baseline models.
The Algonauts Project 2025 Challenge~\citep{gifford2024algonauts} turns CNeuroMod movie fMRI data into an encoding benchmark with private in-distribution and out-of-distribution test sets, public leaderboards, and a baseline model.
Similar releases exist for other recording modalities: LibriBrain~\citep{ozdogan2025libribrain} provides 52 hours from one subject, standardized splits, public code, and baseline speech-decoding tasks.
Together, such datasets support model comparison and progress tracking.

\textbf{Remaining gap}
fMRI provides whole-brain spatial coverage and a better decoding-performance-to-stimulus-duration ratio than M/EEG~\citep{banville_scaling_2025}.
However, as summarized in Table~\ref{table:related_datasets}, brain decoding still lacks fMRI datasets that are both deep at the subject level and released as ready-to-use benchmarks.
Such datasets should provide aligned stimulus representations, held-out evaluation sessions, standardized metrics, and reproducible baseline decoders.

\textbf{Contributions}
We introduce \dataset, a deep per-subject dataset for benchmarking decoding models of naturalistic speech from fMRI.
The main contributions are:
\begin{itemize}[itemsep=0em,topsep=0em]
    \item \textbf{A deep fMRI dataset} whose current release comprises 30 hours per subject of whole-brain 3T fMRI from 3 subjects listening to naturalistic speech;
    \item \textbf{Paired brain and stimulus data} linking preprocessed fMRI activity to podcast audio, transcripts, and derived stimulus embeddings;
    \item \textbf{A brain decoding benchmark} for retrieving podcast audio segments from held-out fMRI activity, with standardized splits, metrics, and reproducible baselines;
    \item \textbf{A decoding scaling analysis} showing that performance improves approximately linearly on a log--log scale with per-subject data, with no saturation over the observed range.
\end{itemize}

\section{The \dataset dataset}

We release preprocessed fMRI recordings paired with stimulus representations to support the development and benchmarking of brain decoding and encoding models.
The dataset and baseline code are hosted at \href{https://huggingface.co/datasets/karavela/cephalonauts_one}{\texttt{huggingface.co/datasets/karavela/cephalonauts\_one}}, where the repository README provides additional loading instructions, decoding baselines, and usage details.
After authentication and access approval, the data can be loaded with the Hugging Face \texttt{datasets} package:

\iflatexml
\begin{verbatim}
from datasets import load_dataset

dataset_name = "karavela/cephalonauts_one"
subset_name = "sub-0_fsaverage5"
dataset = load_dataset(dataset_name, name=subset_name,
                       split="train", streaming=True)
row = next(iter(dataset))

print(list(row))
# ["timestamps", "text", "X", "Y_text", "Y_audio", "sub", "ses", "run",
#  "tr", "fmri_offset", "audio", "audio_offset", "audio_duration",
#  "run_fmri_duration", "fmri", "split"]

print("fMRI dim:", len(row["X"]),
      "audio embedding dim:", len(row["Y_audio"]))
# fMRI dim: 20484, audio embedding dim: 3584
\end{verbatim}
\else
\begin{lstlisting}[style=pythoncode]
from datasets import load_dataset

dataset_name = "karavela/cephalonauts_one"
subset_name = "sub-0_fsaverage5"
dataset = load_dataset(dataset_name, name=subset_name,
                       split="train", streaming=True)
row = next(iter(dataset))

print(list(row))
# ['timestamps', 'text', 'X', 'Y_text', 'Y_audio', 'sub', 'ses', 'run',
#  'tr', 'fmri_offset', 'audio', 'audio_offset', 'audio_duration',
#  'run_fmri_duration', 'fmri', 'split']

print(f"fMRI dim: {len(row['X'])}, audio embedding dim: {len(row['Y_audio'])}")
# fMRI dim: 20484, audio embedding dim: 3584
\end{lstlisting}
\fi

The fMRI data are provided in multiple spatial representations, summarized in Table~\ref{table:available_brain_data}, including both volumetric and cortical surface representations, each released on Hugging Face as a separate subject--space subset.
We also provide a separate \texttt{timeseries} subset with run-level preprocessed time series, masks, transcripts, and audio files, allowing users to rebuild the aligned data or define new analyses.
The stimulus data are likewise released in multiple embedding formats derived from both text and audio, as detailed in Table~\ref{table:available_stimulus}.

\subsection{Dataset overview}

\iflatexml
\begin{table}
\else
\begin{wraptable}{r}{0.58\textwidth}
\fi
    \caption{\textbf{Dataset size by subject.}}
    \label{tab:subject_dataset_size}
    \centering
    \setlength{\tabcolsep}{4pt}
    \renewcommand{\arraystretch}{1.15}
    \begin{tabular}{lrrrr}
      \hline
      \rowcolor{slate100}
      \textbf{Subject} & \textbf{\# TRs} & \makecell{\textbf{\#} \textbf{sessions}} & \makecell{\textbf{\#} \textbf{runs}} & \makecell{\textbf{Duration} \textbf{(h)}} \\
      \hline
      sub-0 & \papernum{207283}{207,283} & 24 & 120 & 29.94 \\
      sub-1 & \papernum{209366}{209,366} & 24 & 121 & 30.24 \\
      sub-2 & \papernum{201647}{201,647} & 24 & 117 & 29.13 \\
      \hline
      \rowcolor{indigo50}
      \textbf{Total} & {\bfseries \papernum{618296}{618,296}} & \textbf{72} & \textbf{358} & \textbf{89.31} \\
      \hline
    \end{tabular}
\iflatexml
\end{table}
\else
\end{wraptable}
\fi
\textbf{Acquisition protocol}
As illustrated in Figure~\ref{fig:dataset_overview}, \dataset consists of fMRI recordings from three subjects while they listened to podcasts in their native language, French.
Each session comprised approximately 75 minutes of functional scanning, together with one anatomical T1-weighted scan.
The dataset contains 30 hours of functional recordings per subject (see Table~\ref{tab:subject_dataset_size}).
Within each subject, stimuli were never repeated, although some stimuli were shared across subjects.
Each functional session consisted of five 15-minute runs organized around three distinct podcasts: runs 1--2 and 3--4 presented two podcasts split across consecutive runs, while run 5 presented a third podcast.
For each run, the timing of audio onset relative to MRI acquisition onset was recorded to synchronize the stimulus with the fMRI time series.
Participants were instructed to listen attentively to each podcast run.
After each run, they answered three multiple-choice comprehension questions about the story they had just heard.
Each question had three candidate answers, from which the subject selected one.
These questions were generated automatically using Gemini 3 Pro \citep{team_gemini_2024}.
Three healthy adult volunteers, aged 31--51 years, were recruited for this study.
All participants were native French speakers, had no history of neurological or psychiatric disorders, and met standard MRI safety criteria.
The study was approved by an ethics committee (CPP Ile de France 2 2025-A02302-47 on February 16, 2026), with written informed consent given by all participants and monetary compensation provided (80\,\texteuro{} per session).

\textbf{Stimuli}
Subjects listened to \textit{Les Pieds sur Terre}\footnote{\url{https://www.franceculture.fr/emissions/les-pieds-sur-terre}}, a popular French podcast produced by the public radio station France Culture that features real-life stories told by ordinary people.
Episodes are approximately 30 minutes long and center on first-person true stories about themes such as everyday life, relationships, work, personal turning points, or social change.
Each episode typically features a single main speaker, who narrates a personal experience in rich and diverse language.
Although the same journalist generally introduces the episodes, the story itself is usually delivered as a monologue by the featured speaker.
The narration is often interspersed with short musical interludes.

\subsection{Brain activity recordings}
\label{section:brain_recordings}

\begin{table}[t]
  \caption{
    \textbf{Released brain data.}
    We release three versions of the preprocessed fMRI data generated with \texttt{DeepPrep}: native \texttt{T1w} space, \texttt{MNI} space, and the \texttt{fsaverage5} cortical surface.
    $V$ denotes the number of voxels or surface vertices.
  }
  \label{table:available_brain_data}
  \centering
  \setlength{\tabcolsep}{4pt}
  \renewcommand{\arraystretch}{1.1}
  \begin{tabular}{lcccccp{5cm}}
    \hline
    \rowcolor{slate100}
    \textbf{Space} & \textbf{Vol.} & \textbf{Surf.} & \makecell{\textbf{Subject} \\ \textbf{anat.}} & \makecell{\textbf{Template} \\ \textbf{anat.}} & $V$ & \textbf{Description} \\
    \hline
    \texttt{T1w} & \cmark & \xmark & \cmark & \xmark & $\sim$\papernum{50000}{50,000} &
    whole-brain in each subject's native anatomical space \\
    \texttt{MNI} & \cmark & \xmark & \xmark & \cmark & \papernum{69765}{69,765} &
    whole-brain normalized to 3\,mm \texttt{MNI152NLin6Asym} standard space \\
    \texttt{fsaverage5} & \xmark & \cmark & \xmark & \cmark & \papernum{20484}{20,484} &
    both hemispheres on the standard \texttt{fsaverage5} surface \\
    \hline
  \end{tabular}
\end{table}

\textbf{MRI acquisition}
Data were acquired at the Paris Brain Institute on a 3T Siemens MAGNETOM Cima.X whole-body MRI scanner equipped with a 64-channel head/neck coil.
Functional images were collected with a highly accelerated 2D gradient-echo echo-planar imaging (EPI) sequence using simultaneous multi-slice acquisition with multiband factor 6, yielding a repetition time (TR) of 520\,ms.
The protocol was designed to provide high temporal resolution while maintaining whole-brain coverage.
Functional images were acquired at 3\,mm isotropic resolution.
Each run contained $\sim$1700 functional brain volumes.
High-resolution anatomical images were acquired with a 3D T1-weighted spoiled gradient-echo Dixon sequence at 0.8\,mm isotropic resolution.
The same acquisition protocol was used across all sessions and subjects.
Each subject underwent 30 hours of fMRI scanning, yielding $\sim$207,000 brain volumes.
Additional sequence details are provided in Appendix~\ref{appendix:acquisition}, including the core acquisition parameters summarized in Table~\ref{tab:mri_core_params}.

\textbf{Preprocessing}
Functional images were preprocessed with \texttt{DeepPrep}~\citep{ren_deepprep_2025}, a fully automated fMRI preprocessing pipeline that includes standard spatial preprocessing steps such as motion correction, spatial normalization, and anatomical-to-functional alignment.
\texttt{DeepPrep} uses \texttt{FreeSurfer}~\citep{fischl_freesurfer_2012} for anatomical processing, including cortical surface reconstruction and segmentation.
As summarized in Table~\ref{table:available_brain_data}, we release the resulting data in three complementary spaces: subject-native \texttt{T1w} space, \texttt{MNI152NLin6Asym} space, and the \texttt{fsaverage5} cortical surface.
These representations support different use cases: subject-native \texttt{T1w} space preserves individual anatomy for within-subject analyses, \texttt{MNI} space facilitates cross-subject comparisons and compatibility with standard atlases, and \texttt{fsaverage5} focuses the input on cortical gray matter while reducing dimensionality.
Broad cortical representations are particularly relevant for natural speech fMRI, since semantic responses during story listening are distributed across large regions of cortex \citep{huth_natural_2016}.
Surface projections were computed from each subject's anatomical scan to ensure anatomically consistent mapping from volumetric data to cortex.
For the volumetric \texttt{T1w} and \texttt{MNI} spaces, we apply the corresponding brain mask before export, so each run is stored as a lighter and more decoding-ready two-dimensional time-by-voxel array rather than as a 4D image; the masks are also released to map these arrays back to brain space.
For decoding analyses, each voxel or vertex time series is additionally detrended, z-scored, and temporally smoothed with a moving-average filter of width 4 TRs.

\subsection{Stimulus}
\label{section:stimulus}

\textbf{Stimulus preprocessing}
For each run, we release the original stimulus in two modalities: the raw audio signal and its transcript.
Word-level transcripts were generated automatically with AssemblyAI Universal 3 Pro.
We converted the word-level transcript into a sequence of regularly sampled stimulus onsets spaced every 4 TRs, corresponding to 2.08 seconds.
At each onset, we extracted the immediately preceding context by collecting the last 20 completed words whose offsets occurred before that time, and used their concatenation as the text stimulus.
For the audio modality, we used the same retained onsets and represented each event by the 10-second audio segment ending at that onset.
Early onsets with insufficient preceding context were discarded.
As summarized in Table~\ref{table:available_stimulus}, we release both these segment-level data and the corresponding embeddings.

\textbf{Stimulus embeddings}
We embedded text segments with \texttt{Qwen3-Embedding-8B} and audio segments with \texttt{LCO-Embedding-Omni-7B}, yielding 4096-dimensional and 3584-dimensional representations, respectively~\citep{zhang_qwen3_2025,xiao2025scaling}.
This choice is motivated by the fact that they are leading models on their respective benchmarks: MTEB multilingual for text~\citep{muennighoff2023mteb} and MAEB for audio~\citep{assadi2026maeb}.
These benchmarks evaluate general-purpose embedding spaces across diverse retrieval and similarity tasks, matching our use of text and audio embeddings as targets for held-out segment retrieval.

\textbf{Alignment to fMRI}
Each retained stimulus event was paired with the corresponding fMRI measurement by selecting the brain frame acquired after the event, accounting for the hemodynamic lag, set to 2.5 seconds.
This fixed lag approximates the delay between stimulus processing and the measured BOLD response.
This procedure yields aligned pairs of brain volumes and stimulus embeddings for any machine learning tasks, as illustrated in Figure \ref{fig:decoding_framework}.A.

\begin{table}[t]
  \centering
  \caption{
    \textbf{Released stimulus representations.}
    For each stimulus modality, we release both segment-level data and embeddings from foundation models.
    $D$ denotes the embedding dimensionality.
  }
  \label{table:available_stimulus}
  \setlength{\tabcolsep}{9pt}
  \renewcommand{\arraystretch}{1.1}
  \begin{tabular}{llcp{7.15cm}}
    \hline
    \rowcolor{slate100}
    \rule{0pt}{2.2ex}\textbf{Modality} & \textbf{Representation} & $D$ & \textbf{Description} \\
    \hline
    \texttt{Text} & segments & -- & overlapping 20-word transcript contexts sampled at retained onsets \\
                  & embeddings & \papernum{4096}{4,096} & embeddings of transcript segments computed with \texttt{Qwen3-Embedding-8B} \\
    \hline
    \texttt{Audio} & segments & -- & 10-second audio segments ending at retained onsets \\
                   & embeddings & \papernum{3584}{3,584} & embeddings of audio segments computed with \texttt{LCO-Embedding-Omni-7B}\\
    \hline
  \end{tabular}
\end{table}

\section{Decoding experiments}

\begin{figure}[t]
  \centering
  \iflatexml
    \includegraphics[width=800pt]{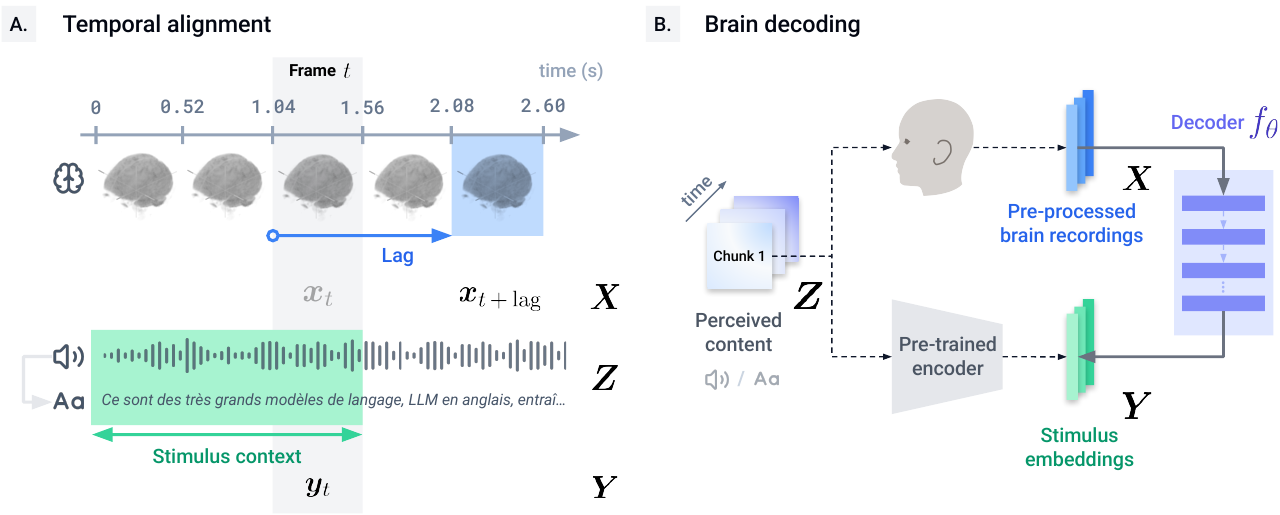}
  \else
    \makebox[\textwidth][c]{\includegraphics[width=1.0\textwidth]{figures/decoding.pdf}}
  \fi
    \caption{
      \textbf{Dataset construction and decoding framework.}
      \textbf{A.}
      For each frame $t$, stimulus $\vz_t$ is aligned with fMRI recordings sampled every 520 ms.
      Target stimulus embeddings $\vy_t$ summarize the audio waveform or transcript over the \textit{stimulus context} leading up to $t$.
      The corresponding brain input uses preprocessed fMRI volumes shifted by a fixed \textit{lag}.
      In this illustration, the \textit{stimulus context} spans 1.56 seconds (3 TRs) and the \textit{lag} is set to 1.04 seconds (2 TRs).
      In the experiments, we set the lag to 2.5 seconds and the stimulus context to 10 seconds.
      \textbf{B.}
      A pretrained encoder maps stimulus $\mZ$ to stimulus embeddings $\mY$.
      The decoder $f_\theta$ is trained to predict $\mY$ from pre-processed brain recordings $\mX$.
    }
    \label{fig:decoding_framework}
\end{figure}

\dataset provides a benchmark for decoding naturalistic speech from fMRI.
The task is formulated as audio segment retrieval: given brain activity from a held-out session, the model must identify the corresponding podcast audio segment among all other possible segments from that session.
This task fixes the evaluation target while leaving the decoding approach open: models may use any stimuli representation, decoder architecture, or training objective, provided they are evaluated on the same held-out retrieval problem.
The proposed baseline predicts embeddings of the stimulus' raw audio and transcript from fMRI activity and ranks candidate segments by cosine similarity in that embedding space.
Figure~\ref{fig:decoding_framework} presents the overall framework.

\subsection{Experimental setup}

\textbf{Decoding task}
For each subject, we evaluate a subject-specific decoder on five folds, using session indices 19--23 once each for test.
Each fold holds out the next session in this set for validation (wrapping from 23 to 19); the remaining 22 sessions form the training pool.
This split ensures that validation and test samples come from recording sessions and podcast segments that were not used for training to more accurately measure generalization.
At test time, the decoder receives an fMRI frame and has to retrieve the corresponding podcast audio segment among approximately $N=2,000$ candidates from the held-out session.
The candidates therefore come from unseen podcast segments in the same session, making retrieval difficult because many segments share the same speaker, topic, and context.

\textbf{Evaluation metrics}
For each test sample $i$, the decoder predicts an embedding $\hat{\vy}_i$ from fMRI activity.
The $N$ stimulus embeddings from the held-out session, $\{\vy_j\}_{j=1}^{N}$, are used as retrieval candidates.
We compute cosine similarities $s_{ij}$ between the predicted embedding and each candidate, and define $r_i$ as the rank of the true aligned stimulus, i.e. how many other candidates are closer to the prediction than the true stimulus is:
\begin{equation}
    s_{ij} \; \triangleq \; \frac{\hat{\vy}_i^\top \vy_j}{\|\hat{\vy}_i\|_2 \; \|\vy_j\|_2},
    \qquad
    r_i \; \triangleq \; \sum_{j=1}^{N} \mathbf{1}\left[s_{ij} > s_{ii}\right] + 1,
\end{equation}
where $\mathbf{1}[\cdot]$ is 1 when the condition holds, 0 otherwise.
We report three metrics.
\textit{Top-10@N} is the fraction of test samples whose true stimulus is among the top 10 candidates:
\begin{equation}
    \text{Top-10@N} \; \triangleq \; \frac{1}{N}\sum_{i=1}^{N} \mathbf{1}\left[r_i \leq 10\right].
\end{equation}
\textit{Median Relative Rank} measures normalized median rank of the true stimulus:
\begin{equation}
    \text{Median Relative Rank} \; \triangleq \; \operatorname{median} \left(\left\{\frac{r_i}{N} \mid i \in \{1,\ldots,N\}\right\}\right).
\end{equation}
Finally, \textit{Cosine Similarity} measures the similarity between the predicted embedding and the true stimulus embedding, averaged across test samples:
\begin{equation}
    \text{Cosine Similarity} \; \triangleq \; \frac{1}{N}\sum_{i=1}^{N} s_{ii}.
\end{equation}
These metrics capture complementary aspects of decoding performance.
Top-10 Retrieval Accuracy measures whether the decoder places the correct stimulus among a small set of plausible candidates.
Median Relative Rank summarizes the ranking distribution and is not sensitive to a single arbitrary cutoff nor to the retrieval set size.
Cosine Similarity measures the quality of the predicted embedding directly, independently of the candidate pool used for retrieval.
Together, these metrics distinguish accurate embedding prediction from successful retrieval among held-out stimulus segments.

\textbf{Baseline decoder}
We provide a simple decoder baseline for this retrieval task.
The decoder is a one-hidden-layer neural network trained to predict a stimulus embedding from brain activity.
Each target embedding is defined as the concatenation of the text and audio embeddings described in Section~\ref{section:stimulus}.
We train the decoder with AdamW~\citep{kingma_adam_2017, loshchilov2017decoupled} using the CLIP contrastive loss~\citep{radford_learning_2021} with temperature 0.02, following recent brain decoding setups~\citep{scotti_reconstructing_2023,jalouzot_optimizing_2025}.
Training uses early stopping on the validation Median Relative Rank, with patience of five epochs, and the best validation checkpoint is used for test evaluation.
At evaluation time, candidate segments are ranked by cosine similarity between their embeddings and the predicted embedding.
All hyperparameters are reported in Appendix~\ref{appendix:hyperparameters}.

\subsection{Results}
\label{section:results}

\begin{figure}[t]
  \centering
  \iflatexml
    \lxAddClass{ltx_flex_cell}
    \includegraphics[width=800pt]{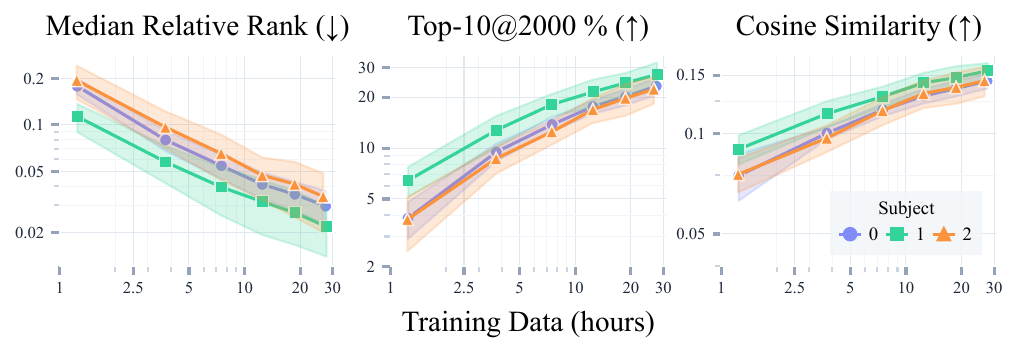}
  \else
    \makebox[\textwidth][c]{\includegraphics[width=1.0\textwidth]{figures/main_scaling_laws_V2.pdf}}
  \fi
  \caption{
    \textbf{Speech decoding performance versus training-set size.}
      Test performance is shown versus training hours for each subject, using Median Relative Rank, Top-10@2000, and Cosine Similarity.
      Lower is better for Median Relative Rank; higher is better for the other metrics.
      Error bars show one empirical standard deviation across five folds with distinct held-out test sessions.
      Performance improves with additional subject-specific data, with no clear saturation.
      Chance level is 0.5 for Median Relative Rank and $\sim$0.5$\%$ for Top-10@2000.
  }
  \label{fig:scaling_laws}
\end{figure}

\textbf{Performance scales with training data}
Figure~\ref{fig:scaling_laws} shows decoding performance as a function of training-set size.
Across subjects, performance improves as more subject-specific training sessions are added.
Cosine Similarity increases, Median Relative Rank decreases, and Top-10 Retrieval Accuracy increases over the range of training sizes.
For example, Cosine Similarity increases from approximately 0.075--0.090 at the smallest training sizes to approximately 0.144--0.155 at the largest training sizes.
Over the same range, Median Relative Rank decreases from approximately 0.11--0.19 to 0.022--0.034, while Top-10 Retrieval Accuracy increases from approximately 3.8--6.4\% to 22.3--27.3\%.
These gains are visible across all three metrics, indicating that the decoder consistently benefits from additional within-subject recordings.
As in the empirical scaling laws reported for neural language models~\citep{kaplan_scaling_2020}, the trajectories are approximately linear in the log-log space of Figure~\ref{fig:scaling_laws}, with performance continuing to improve up to the largest training sets and no clear saturation over the range studied.
This supports the value of deep per-subject datasets for naturalistic language decoding.

\section{Discussion}

\dataset is designed as both a neuroscience resource and a machine-learning benchmark for naturalistic language decoding.
Its main contribution is to increase within-subject depth for speech fMRI, while pairing brain activity with aligned text and audio representations.
This design creates opportunities for studying language processing, evaluating multimodal decoding models, and developing methods that may later transfer to clinical settings.
At the same time, the dataset has important limitations: it includes only three subjects, focuses on French podcast stimuli, and evaluates decoding under a controlled retrieval setting.
We discuss these broader impacts and limitations below.

\subsection{Broader impact}

\textbf{Neuroscience}
\dataset can support a better understanding of how the human brain processes naturalistic speech.
By pairing many hours of fMRI with aligned audio and transcript representations, the dataset makes it possible to study language comprehension over extended timescales, from acoustic input to semantic content.
Its depth also supports subject-specific analyses.

\textbf{Multimodal machine learning}
\dataset also provides a useful setting for multimodal representation learning.
Text and audio benefit from strong pretrained foundation models, whereas fMRI does not yet have comparably mature foundation models.
This creates an asymmetric multimodal problem: one modality has rich pretrained representations, while the other remains noisy, high-dimensional, and data-limited.
The benchmark therefore supports the training and evaluation of models that align brain activity with established text and audio embedding spaces.
More broadly, it can help study how machine learning systems handle modalities with unequal levels of representation quality, data availability, and pretrained model support.

\textbf{Potential negative impacts}
Brain recordings are sensitive biomedical data, and decoding cognition could raise privacy concerns.
To best protect participants privacy, access to \dataset is gated on Hugging Face and requires users to accept terms of use before requesting access.
Participants gave informed consent for data sharing, and the terms of use prohibit re-identification attempts, contact with study volunteers, and provision of information linking the data to protected medical information.

\subsection{Limitations}

\textbf{Limited number of subjects}
\dataset prioritizes depth over breadth.
It contains many hours of recordings per subject, but only three subjects.
This design supports subject-specific decoding and scaling analyses, but limits conclusions about population-level variability and cross-subject generalization.

\textbf{Single-site acquisition}
All data were collected at the same imaging site, on the same scanner, with the same MRI sequence.
This improves consistency across sessions and subjects, but limits conclusions about robustness to other sites, scanners, field strengths, and acquisition settings.

\textbf{MRI acquisition sequence}
The acquisition protocol was optimized for temporal sampling rather than spatial resolution.
The 520\,ms TR provides many brain volumes per hour of listening.
This choice came with a 3\,mm isotropic functional resolution, which is lower than protocols designed for fine-grained spatial mapping.
Thus, \dataset is well suited to studying temporally dense naturalistic speech decoding, but it is less suited to questions that require precise localization of small cortical structures.

\textbf{Restricted language and stimulus domain}
All subjects listened to French podcasts from a single program, \textit{Les Pieds sur Terre}.
This gives coherent and naturalistic stimuli, but restricts the dataset to French narrative speech.
Models trained on \dataset may not generalize to other languages or broader auditory environments.

\textbf{Retrieval-based evaluation}
Our benchmark focuses on retrieval: the model must identify the correct audio segment from a finite pool of candidates in a held-out session.
This setting is already challenging because candidates come from unseen podcast segments with similar topic, speaker, and narrative context.
However, it does not cover the full range of possible decoding tasks.
We hope that \dataset will also support future benchmarks at different levels of difficulty, from easier tasks such as speech-versus-non-speech or coarse stimulus classification to harder tasks such as open-ended generation of text or audio stimuli from brain activity.

\section{Conclusion}

\dataset provides 30 hours of whole-brain 3T fMRI per subject from three subjects listening to French podcasts, paired with aligned audio and transcript representations.
It includes preprocessed brain data in multiple spaces, stimulus embeddings from text and audio, standardized decoding tasks, baseline models, and scaling analyses over training duration.
Decoding performance improves with additional within-subject data, with no clear saturation over the range studied, supporting deep recordings for naturalistic language decoding.

\newpage
\begin{ack}
The authors thank Julie Dessoubrais, Brunhilde Ventre, Margaux Couppey, Chahrazad Chenaf, Marine Kavalko, Marjorie Notolan, Silvia Adins, Stéphanie Anastacio, Marie Ropiquet, Jasmina Wallace, Marc Lapert, Benoît Béranger, Clément Barbe, Mélanie Didier, Éric Bardinet, and Stéphane Lehéricy for their help with data collection.

This work was funded by Karavela and benefited from HPC resources provided by GENCI--IDRIS (Grant 2025-AD010316590R1).
The authors declare no additional competing interests beyond their listed primary affiliations.
\end{ack}

\bibliographystyle{plainnat}
\bibliography{bibs/zotero_alexis_thual,bibs/main}

\newpage
\appendix
\setcounter{section}{0}
\setcounter{figure}{0}
\setcounter{table}{0}
\renewcommand{\thesection}{S\arabic{section}}
\renewcommand{\thefigure}{S\arabic{figure}}
\renewcommand{\thetable}{S\arabic{table}}
\renewcommand{\theHfigure}{S\arabic{figure}}
\renewcommand{\theHtable}{S\arabic{table}}
\renewcommand{\thesubsection}{\thesection.\arabic{subsection}}
\renewcommand{\thesubsubsection}{\thesection.\arabic{subsection}.\arabic{subsubsection}}

\section{A visual glimpse at the data}

\begin{figure}[ht]
  \centering
  \iflatexml
    \includegraphics[width=800pt]{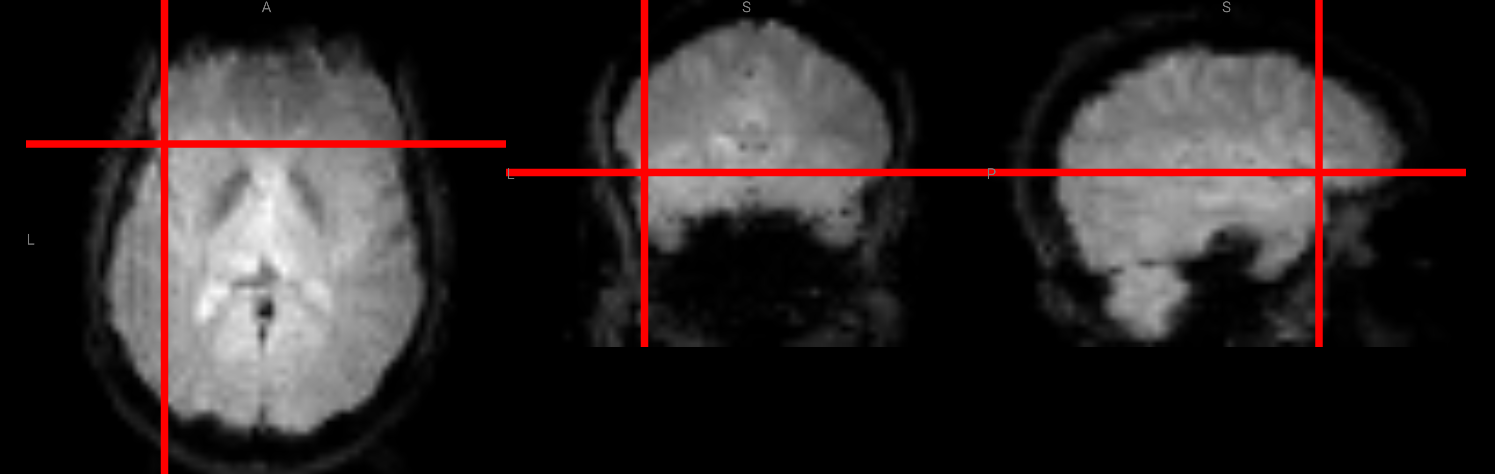}
  \else
    \makebox[\textwidth][c]{\includegraphics[width=1.4\textwidth]{figures/brain_data_frame.png}}
  \fi
  \caption{
    \textbf{Single brain data frame from \dataset.}
    This image is a screenshot obtained with \texttt{niivue} (\url{https://niivue.com/}) of a single brain volume from \dataset, after preprocessing and before masking.
    It shows all three views of the brain (axial, coronal and sagittal) in the native anatomical space of one participant.
  }
  \label{fig:brain_data_frame}
\end{figure}

Figure \ref{fig:brain_data_frame} shows a single brain data frame (or TR) from \dataset, after preprocessing and before masking.
To the expert as well as to the untrained eye, this shows that brain recordings are typically noisy, and may contain reconstruction artifacts.
However, as demonstrated in the decoding results of Section~\ref{section:results}, these data still contain meaningful information, and decoders can successfully learn to leverage it to predict precise information about the perceptual experience of the participants.

\subsection{Interactive exploration notebook}
\label{appendix:colab}

To make the dataset easier to inspect, we provide an interactive Google Colab notebook:
\href{https://colab.research.google.com/drive/1YTF_yOmYMuM8Fg69BTjRaGwIQvrBU3U1?usp=sharing}{\texttt{colab.research.google.com/drive/1YTF\_yOmYMuM8Fg69BTjRaGwIQvrBU3U1}}.
The notebook is a lightweight companion to the Hugging Face repository and the baseline code.
It guides users through authenticating to Hugging Face, loading one validation Parquet file with the \texttt{datasets} library, inspecting the available fields, and visualizing example brain and stimulus data.
The example corresponds to the validation split for subject 1 in the \texttt{fsaverage5} preprocessing space, and includes the aligned fMRI vector, text and audio embeddings, transcript context, session and run identifiers, TR index, timing offsets, and source audio and fMRI paths.
It is intended for quick exploration in a browser, without requiring a local installation.
Full-scale preprocessing, training, and evaluation should still be run from the released code repository.

\subsection{Participant instructions and comprehension questions}
\label{appendix:participant_instructions}

At the beginning of each run, participants received the following instruction:

\begin{mdframed}[backgroundcolor=indigo50, linewidth=0pt, innerbottommargin=0.5cm, innertopmargin=0.5cm, innerleftmargin=0.5cm, innerrightmargin=0.5cm]
  {\sffamily\small
  Vous allez entendre un texte.
  Écoutez-le attentivement, sans bouger.
  Vous pouvez fermer les yeux si vous le souhaitez.
  }
\end{mdframed}

which translates to:

\begin{mdframed}[backgroundcolor=indigo50, linewidth=0pt, innerbottommargin=0.5cm, innertopmargin=0.5cm, innerleftmargin=0.5cm, innerrightmargin=0.5cm]
  {\sffamily\small
  You will hear a text.
  Listen to it attentively, without moving.
  You can close your eyes if you wish.
  }
\end{mdframed}

At the end of each run, participants answered three multiple-choice comprehension questions. For example:

\begin{mdframed}[backgroundcolor=indigo50, linewidth=0pt, innerbottommargin=0.5cm, innertopmargin=0.5cm, innerleftmargin=0.5cm, innerrightmargin=0.5cm]
  {\sffamily\small
  Comment les LLM comme Claude génèrent-ils du texte ?

  \begin{enumerate}[label=\Alph*.]
      \item En analysant le sens profond des mots grâce à une conscience artificielle.
      \item En prédisant token par token (mots, ponctuation) via des calculs de probabilités statistiques.
      \item En recopiant des phrases existantes stockées dans une base de données géante.
  \end{enumerate}
  }
\end{mdframed}

This example question refers to a podcast segment about large language models.
It translates to:

\begin{mdframed}[backgroundcolor=indigo50, linewidth=0pt, innerbottommargin=0.5cm, innertopmargin=0.5cm, innerleftmargin=0.5cm, innerrightmargin=0.5cm]
  {\sffamily\small
  How do LLMs such as Claude generate text?

  \begin{enumerate}[label=\Alph*.]
      \item By analyzing the deep meaning of words through artificial consciousness.
      \item By predicting tokens one by one (words, punctuation) through statistical probability computations.
      \item By copying existing sentences stored in a giant database.
  \end{enumerate}
  }
\end{mdframed}

The correct answer is B.

\subsection{MRI scanner parameters}
\label{appendix:acquisition}

All images were acquired on a 3.0T Siemens MAGNETOM Cima.X whole-body MRI scanner using a 64-channel head/neck coil. Core acquisition parameters for the functional and anatomical scans are summarized in Table~\ref{tab:mri_core_params} while sequence-specific details are provided below.

\begin{table}[ht]
  \centering
  \caption{Core acquisition parameters for the functional and anatomical MRI sequences.}
  \label{tab:mri_core_params}
  \setlength{\tabcolsep}{8pt}
  \renewcommand{\arraystretch}{1.3}
  \begin{tabular}{lll}
    \hline
    \rowcolor{slate100}
    \textbf{Parameter} & \textbf{Functional MRI} & \textbf{Anatomical MRI} \\
    \hline
    Sequence family & 2D gradient-echo EPI & 3D T1-weighted spoiled gradient-echo \\
    Orientation & Transverse & Sagittal \\
    FOV (read $\times$ phase) & $198 \times 198$ mm$^2$ & $230 \times 230$ mm$^2$ \\
    Voxel size & $3.0 \times 3.0 \times 3.0$ mm$^3$ & $0.8 \times 0.8 \times 0.8$ mm$^3$ \\
    Slice thickness & 3.0 mm & 0.8 mm \\
    TR & 520 ms & 6.20 ms \\
    Flip angle & $46^\circ$ & $9^\circ$ \\
    Bandwidth & 2444 Hz/pixel & 480 Hz/pixel \\
    Acquisition time & 15 min 10 s & 3 min 28 s \\
    \hline
  \end{tabular}
\end{table}

\textbf{Functional MRI acquisition}
Functional images were acquired using a 2D gradient-echo echo-planar imaging (EPI) sequence \citep{moeller_multiband_2010} sensitive to blood-oxygen-level-dependent (BOLD) contrast. The acquisition comprised 48 transverse slices with no slice gap, acquired in interleaved order, with an acquisition matrix of $66 \times 66$. Sequence parameters included TE $= 30.0$ ms, posterior-to-anterior phase encoding, and fat saturation. Slice acceleration used simultaneous multi-slice imaging with factor 6, with no in-plane parallel imaging. Other notable parameters were echo spacing $= 0.54$ ms and phase-scrambling was enabled. Each run contained 1736 multi-band volumes together with one single-band reference volume.

\textbf{Anatomical MRI acquisition}
High-resolution anatomical images were acquired using a 3D T1-weighted spoiled gradient-echo Dixon sequence with non-selective excitation. The acquisition consisted of one sagittal slab with 230 slices, reconstructed at isotropic resolution. The readout was Cartesian with bipolar acquisition. The base resolution was 288, with phase resolution $= 100\%$ and slice resolution $= 100\%$. Two echoes were acquired for water--fat separation (TE$_1 = 12.46$ ms, TE$_2 = 23.95$ ms), and images were reconstructed as magnitude volumes. Parallel imaging used a total acceleration factor of 4 ($2 \times 2$, phase-encoding $\times$ partition), with 24 reference lines in each accelerated direction. No partial Fourier or elliptical scanning was used.

\begin{figure}[ht]
  \centering
  \iflatexml
    \includegraphics[width=800pt]{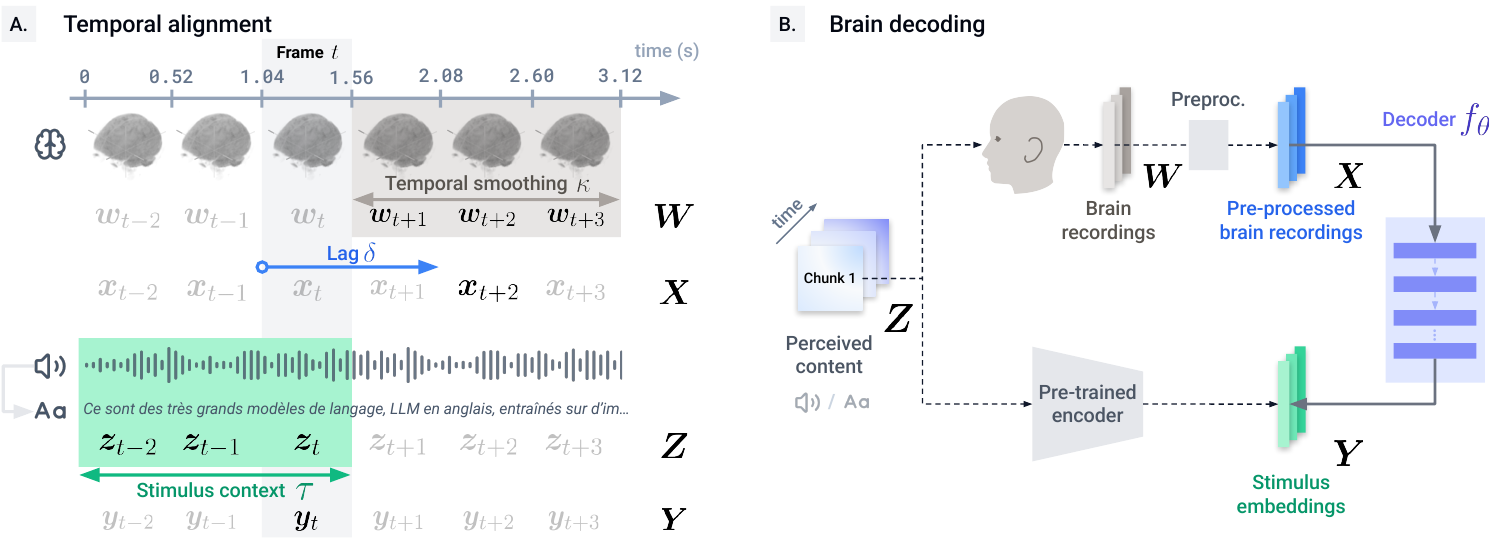}
  \else
    \makebox[\textwidth][c]{\includegraphics[width=1.4\textwidth]{figures/decoding_sup.pdf}}
  \fi
  \caption{
    \textbf{Decoding framework.}
    \textbf{A.}
    Brain activity is represented as a sequence of fMRI volumes sampled every 520 ms (=TR).
    For each frame $t$, we associate a stimulus $\vz_t$ derived from the content perceived while the brain data was recorded during that segment, using either the raw audio or the transcript.
    Then, a stimulus embedding $\vy_t$ is computed over a window of $\tau$ consecutive segments (\textit{stimulus context}).
    The brain data $\vx$ used to decode $\vy_t$ consists of brain volumes $\vw_t$ temporally smoothed over a window of $\kappa$ frames, and taken with a time \textit{lag} $\delta$ with regard to $t$.
    In this illustration, the \textit{temporal smoothing window} $\kappa$ is set to 3 frames -- or TRs, i.e. it is set to 3 $\times$ 520ms -- the \textit{stimulus context} $\tau$ is set to 3 frames, and the \textit{lag} $\delta$ is set to 2 frames.
    \textbf{B.}
    A pretrained encoder produces stimulus embeddings $\mY$ from the stimulus $\mZ$, and a decoder $f_\theta$ is trained to predict these embeddings from the brain recordings $\mX$.
    Decoding can therefore be performed toward text-based or audio-based stimulus representations or even a concatenation of both.\\
    In this figure, variables used to decode stimuli perceived during frame $t$ are shown in black, while variables not used for decoding are shown with reduced opacity.
  }
  \label{fig:decoding_framework_sup}
\end{figure}

\subsection{Formal definition of the brain decoding problem}

For a given subject $p \in P$, a given session $s \in S$ and a given run $r \in R$ in that session, we denote by $\mW_{p, s, r} \in \mathbb{R}^{T_{s, r} \times V}$ the matrix of recorded brain activity during that run, where $T_{s, r}$ is the number of brain volumes acquired during that run and $V$ the number of voxels.\\
Let $\mY_{s, r} \in \mathbb{R}^{T_{s, r} \times D}$ be the matrix of stimulus features derived from the podcast -- either its transcript or the raw audio -- listened to during that run, where $D$ is the dimensionality of the stimulus features.
Multiple types of stimulus features $\mY_{s,r}$ are available, as described in Section \ref{section:stimulus}.\\
Let $\mW_p^\text{train}$ be the concatenation of all brain data recorded for subject $p$ across all sessions and runs that are used for training.
We define $\mY_p^\text{train}$ similarly for the stimulus features, and $\mW_p^\text{val}$, $\mW_p^\text{test}$, $\mY_p^\text{val}$, $\mY_p^\text{test}$ for the validation and test sets.
The set of training, validation and test sessions are disjoint.

\begin{center}
  \begin{minipage}{0.35\textwidth}
    $$
    \mW_p^\text{train} \; \triangleq \; \bigoplus\limits_{\substack{s \in S_\text{train} \\ r \in R}} \mW_{p, s, r} \in \mathbb{R}^{T_\text{train} \times V},
    $$
  \end{minipage}
  \qquad
  \begin{minipage}{0.35\textwidth}
    $$
    \mY_p^\text{train} \; \triangleq \; \bigoplus\limits_{\substack{s \in S_\text{train} \\ r \in R}} \mY_{p, s, r} \in \mathbb{R}^{T_\text{train} \times D},
    $$
  \end{minipage}
  \\
  \begin{minipage}{0.25\textwidth}
    $$
    S \; \triangleq \; S_\text{train} \; \sqcup \; S_\text{val} \; \sqcup \; S_\text{test}
    $$
  \end{minipage}
\end{center}

Let $\mX_p^\text{train}$ be the preprocessed brain data obtained from $\mW_p^\text{train}$ after applying the preprocessing pipeline described in Section~\ref{section:brain_recordings}, and after applying a temporal smoothing function $h$ with window size $\kappa$, such that $\vx_t \in \mathbb{R}^V$ is computed from the $\kappa$ brain volumes $\vw$ centered on $t$, as illustrated in Figure~\ref{fig:decoding_framework_sup}.

The decoding problem consists in learning a function $f_\theta$ parameterized by $\theta$ that maps brain activity to stimulus features, and in evaluating the performance of this function on held-out test data. We refer to $f_\theta$ as the \textit{decoder}.
Since we only train models on data from a single subject, we drop the subscript $p$ for readability.\\
For a given brain volume $t \in T$, the model takes as input the vector of brain activity $\vx_t \in \mathbb{R}^V$ and produces as output a vector of predicted stimulus features $\hat{\vy}_t \triangleq f_\theta(\vx_t) \in \mathbb{R}^D$.
Using gradient descent,  $\theta$ is optimised to minimise a loss function $\mathcal{L}$ that quantifies the discrepancy between the predicted embeddings $\hat{\vy}_t$ and the true embeddings $\vy_t$.

A temporal lag $\delta$ can also be introduced, to account for the delay of the hemodynamic response for instance.
The full model can be written as:
$$
\hat{\vy}_t \; \triangleq \; f_\theta \, (\vx_{t + \delta})
$$

Additionally, the model can be trained on multiple consecutive brain volumes $\vx$, and tasked with predicting stimulus features derived from a window of multiple consecutive stimulus segments, which we did not explore in the present work but which constitute interesting avenues for future research using \dataset.

\section{Preprocessing and decoder training details}
\label{appendix:hyperparameters}

\textbf{Computational resources}
Preprocessing and decoder training were performed on H100-80G GPU workers.
Across all preprocessing and experiments, including preliminary runs not reported in the paper, the total compute was approximately 660 H100-80G GPU-hours.
About 400 GPU-hours were run on cloud computing resources, and about 260 GPU-hours used HPC resources from GENCI--IDRIS (Grant 2025-AD010316590R1).
Most of this compute was used for fMRI preprocessing with DeepPrep.
The baseline decoder itself can also run on consumer hardware, requiring less than 8\,GB of GPU memory and less than 8\,GB of system RAM.

\textbf{Decoder hyperparameters}
Table~\ref{table:decoding_hyperparameters} reports the baseline decoder settings used across the decoding experiments.
These settings define the input fMRI window, hemodynamic lag, stimulus context windows, embedding models, optimization procedure, and retrieval evaluation protocol.

\begin{table}[h]
  \centering
  \caption{
    \textbf{Decoding hyperparameters.}
    Hyperparameters used for all decoding experiments unless otherwise specified.
  }
  \label{table:decoding_hyperparameters}
  \setlength{\tabcolsep}{8pt}
  \renewcommand{\arraystretch}{1.3}
  \begin{tabular}{ll}
    \hline
    \rowcolor{slate100}
    \textbf{Hyperparameter} & \textbf{Value} \\
    \hline
    Decoder architecture & one-hidden-layer MLP \\
    Hidden dimension & 4096 \\
    Activation function & GELU \\
    Dropout & 0.5 \\
    Hemodynamic lag & 2.5 seconds \\
    Text context window & 20 words \\
    Audio context window & 10 seconds \\
    Text embedding model & Qwen3-Embedding-8B \\
    Audio embedding model & LCO-Embedding-Omni-7B \\
    Optimizer & AdamW \\
    Learning rate & 3e-4 \\
    Weight decay & 3e-4 \\
    Batch size & 2048 \\
    Number of epochs & 100 max \\
    Loss function & CLIP-style contrastive loss \\
    Temperature & 2e-2 \\
    Early stopping patience & 5 \\
    Validation criterion & Median Relative Rank \\
    Retrieval set size & $\sim$2000 \\
    Hardware & NVIDIA H100-80G GPU workers \\
    \hline
  \end{tabular}
\end{table}

\begin{figure}[ht]
  \centering
  \iflatexml
    \lxAddClass{ltx_flex_cell}
    \includegraphics[width=800pt]{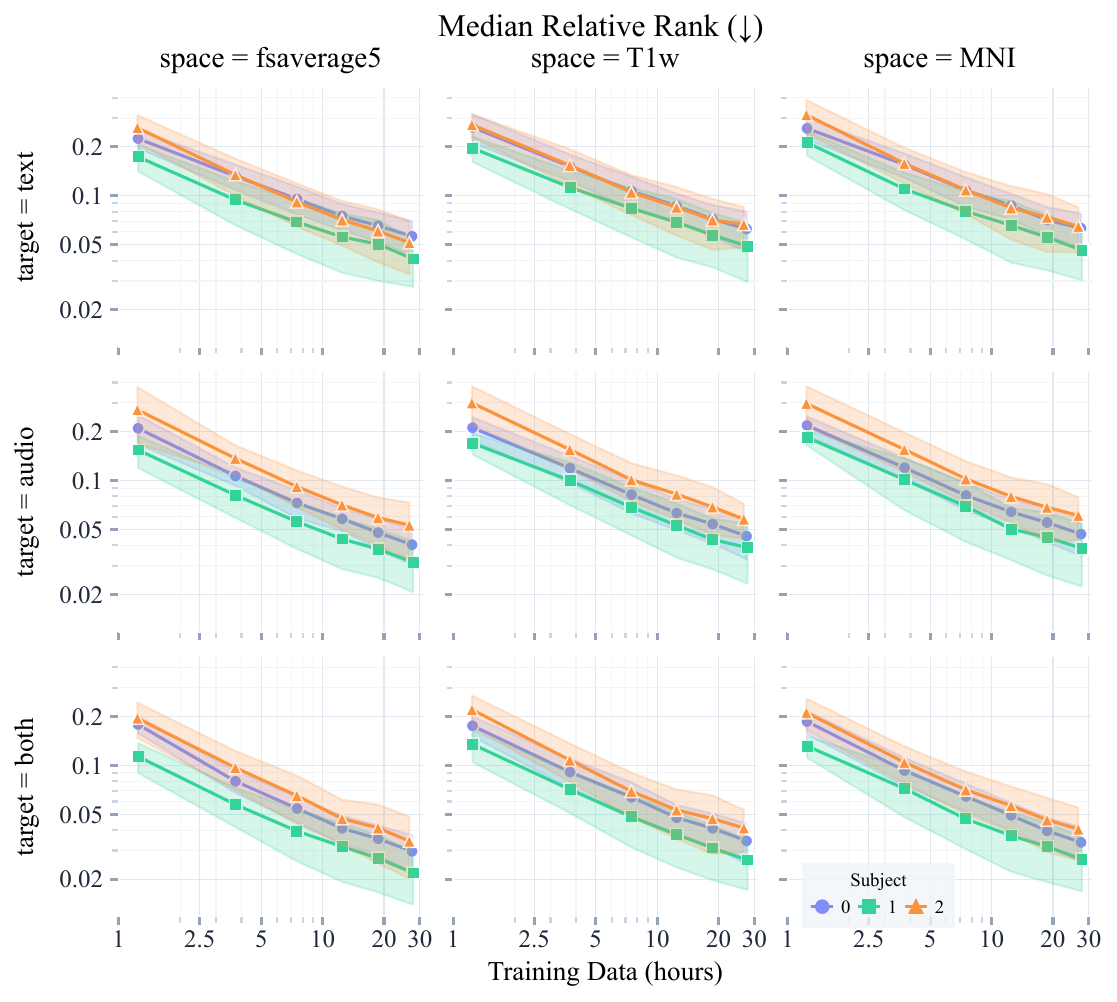}
  \else
    \makebox[\textwidth][c]{\includegraphics[width=1.0\textwidth]{figures/scaling_laws_modalities_V2.pdf}}
  \fi
  \caption{
    \textbf{Supplementary scaling analysis across brain spaces and stimulus targets.}
    Median Relative Rank is shown as a function of subject-specific training hours for each released brain representation and stimulus target.
    The brain representations correspond to the preprocessing outputs listed in Table~\ref{table:available_brain_data}: \texttt{fsaverage5}, \texttt{T1w}, and \texttt{MNI}.
    The stimulus targets correspond to the text, audio, and concatenated embeddings described in Table~\ref{table:available_stimulus}.
    Lower values indicate better retrieval performance.
  }
  \label{fig:scaling_laws_preprocessing_embeddings}
\end{figure}

\textbf{Scaling analysis}
Figure~\ref{fig:scaling_laws_preprocessing_embeddings} extends the main scaling analysis to all released brain preprocessing spaces and stimulus embedding targets.
All combinations provide useful decoding signal: Median Relative Rank improves with additional training data for the \texttt{fsaverage5}, \texttt{T1w}, and \texttt{MNI} representations, and for text, audio, and concatenated embedding targets.
This supports the relevance of the preprocessing outputs released with the dataset, described in Table~\ref{table:available_brain_data}, and of the stimulus embeddings described in Table~\ref{table:available_stimulus}.

\section{Licenses and terms of use for existing assets}
\label{appendix:asset_licenses}

\textbf{Code dependencies}
The released code uses Python, NumPy~\citep{harris2020array}, SciPy~\citep{virtanen_scipy_2020}, PyTorch~\citep{paszke2019pytorch}, torchaudio~\citep{yang2022torchaudio}, Hugging Face datasets~\citep{lhoest2021datasets}, Hugging Face transformers~\citep{wolf_transformers_2020}, Hugging Face Hub tools, scikit-learn~\citep{pedregosa_scikit-learn_2011}, nilearn~\citep{abraham_machine_2014}, sentence-transformers~\citep{reimers2019sentence}, pandas~\citep{mckinney2010data}, matplotlib~\citep{hunter_matplotlib_2007}, seaborn~\citep{waskom_seaborn_2021}, joblib~\citep{joblib_developers_2025}, tqdm~\citep{da_costa_luis_tqdm_2024}, uv, and pip.

\textbf{Embedding models and hosted services}
Transcript embeddings were computed with \texttt{Qwen3-Embedding-8B}~\citep{zhang_qwen3_2025}, released under the Apache License 2.0.
Audio embeddings were computed with \texttt{LCO-Embedding-Omni-7B}~\citep{xiao2025scaling}, also released under the Apache License 2.0.
Transcripts were generated with AssemblyAI Universal 3 Pro; this use was governed by the AssemblyAI Terms of Service.\footnote{\url{https://www.assemblyai.com/legal/terms-of-service}}
Comprehension questions were generated with Gemini 3 Pro~\citep{team_gemini_2024}; this use was governed by the Google APIs Terms of Service and the Gemini API Additional Terms of Service.\footnote{\url{https://ai.google.dev/gemini-api/terms}}
No Gemini model weights are redistributed as part of this work.

\textbf{Preprocessing software}
The fMRI preprocessing used DeepPrep~\citep{ren_deepprep_2025}, which is released under the Apache License 2.0.
DeepPrep uses standard neuroimaging templates distributed through TemplateFlow~\citep{ciric_templateflow_2022}.
DeepPrep relies on FreeSurfer components for anatomical processing~\citep{fischl_freesurfer_2012}; FreeSurfer is distributed under its own software license and requires a free license file for use.\footnote{\url{https://freesurfer.net/fswiki/License}}

\clearpage

\section*{NeurIPS Paper Checklist}

\begin{enumerate}

\item {\bf Claims}
    \item[] Question: Do the main claims made in the abstract and introduction accurately reflect the paper's contributions and scope?
    \item[] Answer: \answerYes{}
    \item[] Justification: The abstract and Introduction state that the paper contributes a deep fMRI dataset, paired audio/transcript representations, a retrieval benchmark, baseline decoders, and a scaling analysis. The Discussion and Limitations sections describe the main scope restrictions, including the small number of subjects, single-site acquisition, French podcast stimuli, and retrieval-based evaluation.
    \item[] Guidelines:
    \begin{itemize}
        \item The answer \answerNA{} means that the abstract and introduction do not include the claims made in the paper.
        \item The abstract and/or introduction should clearly state the claims made, including the contributions made in the paper and important assumptions and limitations. A \answerNo{} or \answerNA{} answer to this question will not be perceived well by the reviewers. 
        \item The claims made should match theoretical and experimental results, and reflect how much the results can be expected to generalize to other settings. 
        \item It is fine to include aspirational goals as motivation as long as it is clear that these goals are not attained by the paper. 
    \end{itemize}

\item {\bf Limitations}
    \item[] Question: Does the paper discuss the limitations of the work performed by the authors?
    \item[] Answer: \answerYes{}
    \item[] Justification: The Discussion contains a dedicated Limitations subsection. It discusses the limited number of subjects, single-site acquisition, restricted language and stimulus domain, and retrieval-based evaluation.
    \item[] Guidelines:
    \begin{itemize}
        \item The answer \answerNA{} means that the paper has no limitation while the answer \answerNo{} means that the paper has limitations, but those are not discussed in the paper. 
        \item The authors are encouraged to create a separate ``Limitations'' section in their paper.
        \item The paper should point out any strong assumptions and how robust the results are to violations of these assumptions (e.g., independence assumptions, noiseless settings, model well-specification, asymptotic approximations only holding locally). The authors should reflect on how these assumptions might be violated in practice and what the implications would be.
        \item The authors should reflect on the scope of the claims made, e.g., if the approach was only tested on a few datasets or with a few runs. In general, empirical results often depend on implicit assumptions, which should be articulated.
        \item The authors should reflect on the factors that influence the performance of the approach. For example, a facial recognition algorithm may perform poorly when image resolution is low or images are taken in low lighting. Or a speech-to-text system might not be used reliably to provide closed captions for online lectures because it fails to handle technical jargon.
        \item The authors should discuss the computational efficiency of the proposed algorithms and how they scale with dataset size.
        \item If applicable, the authors should discuss possible limitations of their approach to address problems of privacy and fairness.
        \item While the authors might fear that complete honesty about limitations might be used by reviewers as grounds for rejection, a worse outcome might be that reviewers discover limitations that aren't acknowledged in the paper. The authors should use their best judgment and recognize that individual actions in favor of transparency play an important role in developing norms that preserve the integrity of the community. Reviewers will be specifically instructed to not penalize honesty concerning limitations.
    \end{itemize}

\item {\bf Theory assumptions and proofs}
    \item[] Question: For each theoretical result, does the paper provide the full set of assumptions and a complete (and correct) proof?
    \item[] Answer: \answerNA{}
    \item[] Justification: The paper introduces a dataset and empirical decoding benchmark. It does not present theoretical results, theorems, or proofs.
    \item[] Guidelines:
    \begin{itemize}
        \item The answer \answerNA{} means that the paper does not include theoretical results. 
        \item All the theorems, formulas, and proofs in the paper should be numbered and cross-referenced.
        \item All assumptions should be clearly stated or referenced in the statement of any theorems.
        \item The proofs can either appear in the main paper or the supplemental material, but if they appear in the supplemental material, the authors are encouraged to provide a short proof sketch to provide intuition. 
        \item Inversely, any informal proof provided in the core of the paper should be complemented by formal proofs provided in appendix or supplemental material.
        \item Theorems and Lemmas that the proof relies upon should be properly referenced. 
    \end{itemize}

    \item {\bf Experimental result reproducibility}
    \item[] Question: Does the paper fully disclose all the information needed to reproduce the main experimental results of the paper to the extent that it affects the main claims and/or conclusions of the paper (regardless of whether the code and data are provided or not)?
    \item[] Answer: \answerYes{}
    \item[] Justification: The paper describes the released data representations, stimulus preprocessing, fMRI alignment, train/validation/test separation, retrieval metrics, baseline decoder, optimizer, loss, and hyperparameters. The data and code release is linked in the abstract and is intended to provide the scripts needed to reproduce the main decoding results.
    \item[] Guidelines:
    \begin{itemize}
        \item The answer \answerNA{} means that the paper does not include experiments.
        \item If the paper includes experiments, a \answerNo{} answer to this question will not be perceived well by the reviewers: Making the paper reproducible is important, regardless of whether the code and data are provided or not.
        \item If the contribution is a dataset and\slash or model, the authors should describe the steps taken to make their results reproducible or verifiable. 
        \item Depending on the contribution, reproducibility can be accomplished in various ways. For example, if the contribution is a novel architecture, describing the architecture fully might suffice, or if the contribution is a specific model and empirical evaluation, it may be necessary to either make it possible for others to replicate the model with the same dataset, or provide access to the model. In general. releasing code and data is often one good way to accomplish this, but reproducibility can also be provided via detailed instructions for how to replicate the results, access to a hosted model (e.g., in the case of a large language model), releasing of a model checkpoint, or other means that are appropriate to the research performed.
        \item While NeurIPS does not require releasing code, the conference does require all submissions to provide some reasonable avenue for reproducibility, which may depend on the nature of the contribution. For example
        \begin{enumerate}
            \item If the contribution is primarily a new algorithm, the paper should make it clear how to reproduce that algorithm.
            \item If the contribution is primarily a new model architecture, the paper should describe the architecture clearly and fully.
            \item If the contribution is a new model (e.g., a large language model), then there should either be a way to access this model for reproducing the results or a way to reproduce the model (e.g., with an open-source dataset or instructions for how to construct the dataset).
            \item We recognize that reproducibility may be tricky in some cases, in which case authors are welcome to describe the particular way they provide for reproducibility. In the case of closed-source models, it may be that access to the model is limited in some way (e.g., to registered users), but it should be possible for other researchers to have some path to reproducing or verifying the results.
        \end{enumerate}
    \end{itemize}

\item {\bf Open access to data and code}
    \item[] Question: Does the paper provide open access to the data and code, with sufficient instructions to faithfully reproduce the main experimental results, as described in supplemental material?
    \item[] Answer: \answerYes{}
    \item[] Justification: The abstract provides a public release link for the dataset and code. The release is intended to include the preprocessed fMRI data, stimulus representations, benchmark splits, metrics, and baseline code needed to reproduce the main results.
    \item[] Guidelines:
    \begin{itemize}
        \item The answer \answerNA{} means that paper does not include experiments requiring code.
        \item Please see the NeurIPS code and data submission guidelines (\url{https://neurips.cc/public/guides/CodeSubmissionPolicy}) for more details.
        \item While we encourage the release of code and data, we understand that this might not be possible, so \answerNo{} is an acceptable answer. Papers cannot be rejected simply for not including code, unless this is central to the contribution (e.g., for a new open-source benchmark).
        \item The instructions should contain the exact command and environment needed to run to reproduce the results. See the NeurIPS code and data submission guidelines (\url{https://neurips.cc/public/guides/CodeSubmissionPolicy}) for more details.
        \item The authors should provide instructions on data access and preparation, including how to access the raw data, preprocessed data, intermediate data, and generated data, etc.
        \item The authors should provide scripts to reproduce all experimental results for the new proposed method and baselines. If only a subset of experiments are reproducible, they should state which ones are omitted from the script and why.
        \item At submission time, to preserve anonymity, the authors should release anonymized versions (if applicable).
        \item Providing as much information as possible in supplemental material (appended to the paper) is recommended, but including URLs to data and code is permitted.
    \end{itemize}

\item {\bf Experimental setting/details}
    \item[] Question: Does the paper specify all the training and test details (e.g., data splits, hyperparameters, how they were chosen, type of optimizer) necessary to understand the results?
    \item[] Answer: \answerYes{}
    \item[] Justification: The Decoding experiments section specifies the held-out-session retrieval task, the use of all segments from the held-out session as retrieval candidates, the metrics, baseline decoder, optimizer, and loss. Appendix~\ref{appendix:hyperparameters} reports the hyperparameters used for the baseline experiments.
    \item[] Guidelines:
    \begin{itemize}
        \item The answer \answerNA{} means that the paper does not include experiments.
        \item The experimental setting should be presented in the core of the paper to a level of detail that is necessary to appreciate the results and make sense of them.
        \item The full details can be provided either with the code, in appendix, or as supplemental material.
    \end{itemize}

\item {\bf Experiment statistical significance}
    \item[] Question: Does the paper report error bars suitably and correctly defined or other appropriate information about the statistical significance of the experiments?
    \item[] Answer: \answerYes{}
    \item[] Justification: Results are reported with error bars across five folds, using session indices 19--23 once each for test. For each fold, the next session in this set is held out for validation (wrapping from 23 to 19), and the remaining 22 sessions form the training pool. The error bars in the scaling-law plots are empirical one-standard-deviation error bars computed by seaborn with \texttt{errorbar="sd"} across the fold results. They therefore capture variability due to the held-out session split, not model initialization. No normality assumption or formal significance test is used; the bars are standard deviations, not standard errors or confidence intervals.
    \item[] Guidelines:
    \begin{itemize}
        \item The answer \answerNA{} means that the paper does not include experiments.
        \item The authors should answer \answerYes{} if the results are accompanied by error bars, confidence intervals, or statistical significance tests, at least for the experiments that support the main claims of the paper.
        \item The factors of variability that the error bars are capturing should be clearly stated (for example, train/test split, initialization, random drawing of some parameter, or overall run with given experimental conditions).
        \item The method for calculating the error bars should be explained (closed form formula, call to a library function, bootstrap, etc.)
        \item The assumptions made should be given (e.g., Normally distributed errors).
        \item It should be clear whether the error bar is the standard deviation or the standard error of the mean.
        \item It is OK to report 1-sigma error bars, but one should state it. The authors should preferably report a 2-sigma error bar than state that they have a 96\% CI, if the hypothesis of Normality of errors is not verified.
        \item For asymmetric distributions, the authors should be careful not to show in tables or figures symmetric error bars that would yield results that are out of range (e.g., negative error rates).
        \item If error bars are reported in tables or plots, the authors should explain in the text how they were calculated and reference the corresponding figures or tables in the text.
    \end{itemize}

\item {\bf Experiments compute resources}
    \item[] Question: For each experiment, does the paper provide sufficient information on the computer resources (type of compute workers, memory, time of execution) needed to reproduce the experiments?
    \item[] Answer: \answerYes{}
    \item[] Justification: The appendix paragraph on computational resources reports the H100-80G GPU workers used for preprocessing and decoder training.
    \item[] Guidelines:
    \begin{itemize}
        \item The answer \answerNA{} means that the paper does not include experiments.
        \item The paper should indicate the type of compute workers CPU or GPU, internal cluster, or cloud provider, including relevant memory and storage.
        \item The paper should provide the amount of compute required for each of the individual experimental runs as well as estimate the total compute. 
        \item The paper should disclose whether the full research project required more compute than the experiments reported in the paper (e.g., preliminary or failed experiments that didn't make it into the paper). 
    \end{itemize}
    
\item {\bf Code of ethics}
    \item[] Question: Does the research conducted in the paper conform, in every respect, with the NeurIPS Code of Ethics \url{https://neurips.cc/public/EthicsGuidelines}?
    \item[] Answer: \answerYes{}
    \item[] Justification: The research conducted in this paper conforms, in every respect, with the NeurIPS Code of Ethics.
    \item[] Guidelines:
    \begin{itemize}
        \item The answer \answerNA{} means that the authors have not reviewed the NeurIPS Code of Ethics.
        \item If the authors answer \answerNo, they should explain the special circumstances that require a deviation from the Code of Ethics.
        \item The authors should make sure to preserve anonymity (e.g., if there is a special consideration due to laws or regulations in their jurisdiction).
    \end{itemize}

\item {\bf Broader impacts}
    \item[] Question: Does the paper discuss both potential positive societal impacts and negative societal impacts of the work performed?
    \item[] Answer: \answerYes{}
    \item[] Justification: The Broader impact section covers positive uses in neuroscience and multimodal representation learning, as well as potential risks such as privacy concerns and misuse of brain decoding results. It also describes the gated access terms used for the dataset release.
    \item[] Guidelines:
    \begin{itemize}
        \item The answer \answerNA{} means that there is no societal impact of the work performed.
        \item If the authors answer \answerNA{} or \answerNo, they should explain why their work has no societal impact or why the paper does not address societal impact.
        \item Examples of negative societal impacts include potential malicious or unintended uses (e.g., disinformation, generating fake profiles, surveillance), fairness considerations (e.g., deployment of technologies that could make decisions that unfairly impact specific groups), privacy considerations, and security considerations.
        \item The conference expects that many papers will be foundational research and not tied to particular applications, let alone deployments. However, if there is a direct path to any negative applications, the authors should point it out. For example, it is legitimate to point out that an improvement in the quality of generative models could be used to generate Deepfakes for disinformation. On the other hand, it is not needed to point out that a generic algorithm for optimizing neural networks could enable people to train models that generate Deepfakes faster.
        \item The authors should consider possible harms that could arise when the technology is being used as intended and functioning correctly, harms that could arise when the technology is being used as intended but gives incorrect results, and harms following from (intentional or unintentional) misuse of the technology.
        \item If there are negative societal impacts, the authors could also discuss possible mitigation strategies (e.g., gated release of models, providing defenses in addition to attacks, mechanisms for monitoring misuse, mechanisms to monitor how a system learns from feedback over time, improving the efficiency and accessibility of ML).
    \end{itemize}
    
\item {\bf Safeguards}
    \item[] Question: Does the paper describe safeguards that have been put in place for responsible release of data or models that have a high risk for misuse (e.g., pre-trained language models, image generators, or scraped datasets)?
    \item[] Answer: \answerYes{}
    \item[] Justification: The Broader impact section states that brain data are sensitive biomedical data and describes the safeguards used for the dataset release. Access to \dataset is gated on Hugging Face, requires acceptance of terms of use, prohibits attempts to identify or contact study volunteers, requires compliance with applicable institutional and ethics-review rules, and requires redistribution under the same conditions.
    \item[] Guidelines:
    \begin{itemize}
        \item The answer \answerNA{} means that the paper poses no such risks.
        \item Released models that have a high risk for misuse or dual-use should be released with necessary safeguards to allow for controlled use of the model, for example by requiring that users adhere to usage guidelines or restrictions to access the model or implementing safety filters. 
        \item Datasets that have been scraped from the Internet could pose safety risks. The authors should describe how they avoided releasing unsafe images.
        \item We recognize that providing effective safeguards is challenging, and many papers do not require this, but we encourage authors to take this into account and make a best faith effort.
    \end{itemize}

\item {\bf Licenses for existing assets}
    \item[] Question: Are the creators or original owners of assets (e.g., code, data, models), used in the paper, properly credited and are the license and terms of use explicitly mentioned and properly respected?
    \item[] Answer: \answerYes{}
    \item[] Justification: Appendix~\ref{appendix:asset_licenses} credits the main code dependencies, embedding models, hosted Gemini service, and preprocessing software used in the work. It also lists applicable licenses or terms of use for the pretrained models and preprocessing tools.
    \item[] Guidelines:
    \begin{itemize}
        \item The answer \answerNA{} means that the paper does not use existing assets.
        \item The authors should cite the original paper that produced the code package or dataset.
        \item The authors should state which version of the asset is used and, if possible, include a URL.
        \item The name of the license (e.g., CC-BY 4.0) should be included for each asset.
        \item For scraped data from a particular source (e.g., website), the copyright and terms of service of that source should be provided.
        \item If assets are released, the license, copyright information, and terms of use in the package should be provided. For popular datasets, \url{paperswithcode.com/datasets} has curated licenses for some datasets. Their licensing guide can help determine the license of a dataset.
        \item For existing datasets that are re-packaged, both the original license and the license of the derived asset (if it has changed) should be provided.
        \item If this information is not available online, the authors are encouraged to reach out to the asset's creators.
    \end{itemize}

\item {\bf New assets}
    \item[] Question: Are new assets introduced in the paper well documented and is the documentation provided alongside the assets?
    \item[] Answer: \answerYes{}
    \item[] Justification: The paper documents the new dataset, released brain and stimulus representations, benchmark task, metrics, baseline model, and hyperparameters. The dataset and code release is intended to provide accompanying documentation for accessing and using these assets.
    \item[] Guidelines:
    \begin{itemize}
        \item The answer \answerNA{} means that the paper does not release new assets.
        \item Researchers should communicate the details of the dataset\slash code\slash model as part of their submissions via structured templates. This includes details about training, license, limitations, etc. 
        \item The paper should discuss whether and how consent was obtained from people whose asset is used.
        \item At submission time, remember to anonymize your assets (if applicable). You can either create an anonymized URL or include an anonymized zip file.
    \end{itemize}

\item {\bf Crowdsourcing and research with human subjects}
    \item[] Question: For crowdsourcing experiments and research with human subjects, does the paper include the full text of instructions given to participants and screenshots, if applicable, as well as details about compensation (if any)? 
    \item[] Answer: \answerYes{}
    \item[] Justification: Appendix~\ref{appendix:participant_instructions} provides the full instruction text given at the beginning of each run and an example multiple-choice comprehension question. The Dataset section reports monetary compensation; screenshots are not applicable to this fMRI listening protocol.
    \item[] Guidelines:
    \begin{itemize}
        \item The answer \answerNA{} means that the paper does not involve crowdsourcing nor research with human subjects.
        \item Including this information in the supplemental material is fine, but if the main contribution of the paper involves human subjects, then as much detail as possible should be included in the main paper. 
        \item According to the NeurIPS Code of Ethics, workers involved in data collection, curation, or other labor should be paid at least the minimum wage in the country of the data collector. 
    \end{itemize}

\item {\bf Institutional review board (IRB) approvals or equivalent for research with human subjects}
    \item[] Question: Does the paper describe potential risks incurred by study participants, whether such risks were disclosed to the subjects, and whether Institutional Review Board (IRB) approvals (or an equivalent approval/review based on the requirements of your country or institution) were obtained?
    \item[] Answer: \answerYes{}
    \item[] Justification: The Dataset section states that participants were healthy adult volunteers who met standard MRI safety criteria, that the study was approved by an ethics committee (CPP Ile de France 2 2025-A02302-47 on February 16, 2026), that all participants gave written informed consent, and that participants received monetary compensation.
    \item[] Guidelines:
    \begin{itemize}
        \item The answer \answerNA{} means that the paper does not involve crowdsourcing nor research with human subjects.
        \item Depending on the country in which research is conducted, IRB approval (or equivalent) may be required for any human subjects research. If you obtained IRB approval, you should clearly state this in the paper. 
        \item We recognize that the procedures for this may vary significantly between institutions and locations, and we expect authors to adhere to the NeurIPS Code of Ethics and the guidelines for their institution. 
        \item For initial submissions, do not include any information that would break anonymity (if applicable), such as the institution conducting the review.
    \end{itemize}

\item {\bf Declaration of LLM usage}
    \item[] Question: Does the paper describe the usage of LLMs if it is an important, original, or non-standard component of the core methods in this research? Note that if the LLM is used only for writing, editing, or formatting purposes and does \emph{not} impact the core methodology, scientific rigor, or originality of the research, declaration is not required.
    \item[] Answer: \answerYes{}
    \item[] Justification: The paper states that Gemini 3 Pro was used to generate comprehension questions, transcript segments were embedded with \texttt{Qwen3-Embedding-8B}, and audio segments were embedded with \texttt{LCO-Embedding-Omni-7B}. These uses are described in the Dataset and Stimulus sections.
    \item[] Guidelines:
    \begin{itemize}
        \item The answer \answerNA{} means that the core method development in this research does not involve LLMs as any important, original, or non-standard components.
        \item Please refer to our LLM policy in the NeurIPS handbook for what should or should not be described.
    \end{itemize}

\end{enumerate}

\end{document}